\documentclass[10pt,twocolumn,letterpaper]{article}

\usepackage{iccv}
\usepackage{times}
\usepackage{epsfig}
\usepackage{graphicx}
\usepackage{amsmath}
\usepackage{amssymb}
\usepackage{gensymb}
\usepackage{xcolor}

\usepackage[breaklinks=true,bookmarks=false]{hyperref}
\usepackage[accsupp]{axessibility} 

\iccvfinalcopy 

\newcommand{\pixelnum}{\mathcal{P}}

\def\iccvPaperID{5300} 

\begin{document}

\title{Towards Viewpoint Robustness in Bird's Eye View Segmentation}




\newcommand{\superscript}[1]{\ensuremath{^{\textrm{#1}}}}
\author{
    Tzofi Klinghoffer\superscript{1,2}\thanks{Work done during an internship at NVIDIA.}\hspace{2mm} 
    Jonah Philion\superscript{2,3,4} \hspace{2mm}   
    Wenzheng Chen\superscript{2,3,4} \hspace{2mm} 
    Or Litany\superscript{2}\\
    Zan Gojcic\superscript{2} \hspace{2mm} 
    Jungseock Joo\superscript{2,5} \hspace{2mm} 
    Ramesh Raskar\superscript{1} \hspace{2mm} 
    Sanja Fidler\superscript{2,3,4} \hspace{2mm} 
    Jose M. Alvarez\superscript{2\textdagger}\\
    \vspace{1mm}
    \normalsize{\superscript{1}MIT \hspace{2mm}
    \superscript{2}NVIDIA \hspace{2mm}
    \superscript{3}University of Toronto \hspace{2mm}
    \superscript{4}Vector Institute \hspace{2mm}
    \superscript{5}UCLA}\\
    \normalsize{\textit{Project page}: \href{https://nvlabs.github.io/viewpoint-robustness/}{https://nvlabs.github.io/viewpoint-robustness}}
}


\maketitle

\def\thefootnote{\textdagger}\footnotetext{Corresponding author: Jose M. Alvarez (josea@nvidia.com).}\def\thefootnote{\arabic{footnote}}


\begin{abstract}
   Autonomous vehicles (AV) require that neural networks used for perception be robust to different viewpoints if they are to be deployed across many types of vehicles without the repeated cost of data collection and labeling for each. AV companies typically focus on collecting data from diverse scenarios and locations, but not camera rig configurations, due to cost. As a result, only a small number of rig variations exist across most fleets. In this paper, we study how AV perception models are affected by changes in camera viewpoint and propose a way to scale them across vehicle types without repeated data collection and labeling. Using bird's eye view (BEV) segmentation as a motivating task, we find through extensive experiments that existing perception models are surprisingly sensitive to changes in camera viewpoint. When trained with data from one camera rig, small changes to pitch, yaw, depth, or height of the camera at inference time lead to large drops in performance. We introduce a technique for novel view synthesis and use it to transform collected data to the viewpoint of target rigs, allowing us to train BEV segmentation models for diverse target rigs without any additional data collection or labeling cost. To analyze the impact of viewpoint changes, we leverage synthetic data to mitigate other gaps (content, ISP, etc). Our approach is then trained on real data and evaluated on synthetic data, enabling evaluation on diverse target rigs. We release all data for use in future work. Our method is able to recover an average of 14.7\% of the IoU that is otherwise lost when deploying to new rigs.

\end{abstract}

\vspace{-3mm}
\section{Introduction}

Neural networks (NNs) are becoming ubiquitous across domains. Safety critical applications, such as autonomous vehicles (AVs), rely on these NN to be robust to out of distribution (OOD) data. Yet, recent work has drawn attention to the susceptibility of NNs to failure when exposed to OOD data, such as adversarial corruptions \cite{hendrycks2019benchmarking}, unseen weather conditions \cite{michaelis2019benchmarking}, and new geographic regions \cite{drenkow2021robustness}. While each of these pose a significant challenge for safety critical applications, we focus on another distribution shift, which, thus far, has been understudied in the research literature -- changes in camera viewpoint between train data and test data. Because camera viewpoint changes are realistic in AVs, we study their impact on AV perception tasks.

AVs use cameras around the ego-vehicle to perceive their surroundings. Using images from each camera, NNs detect and segment objects in the scene, such as vehicles, pedestrians, roads, and more. This information is used by trajectory planners to decide how the ego-vehicle navigates. Camera viewpoint for AVs may differ between train and test in several real-world scenarios. First, the camera viewpoint may change over time due to wear and tear or damage. Second, camera viewpoint may change due to installation variation. Third, and most relevant for our work, if a single NN is to be deployed across different types of vehicles, it must be able to generalize to the camera viewpoints of each car. Collecting and labeling train data for each target rig is not scalable and quickly becomes intractable for AV companies wishing to scale across many types of vehicles due to cost, thus motivating our work to transform collected data into the viewpoint of diverse target rigs to use for training.




\begin{figure}
    \centering
    \includegraphics[width=\linewidth]{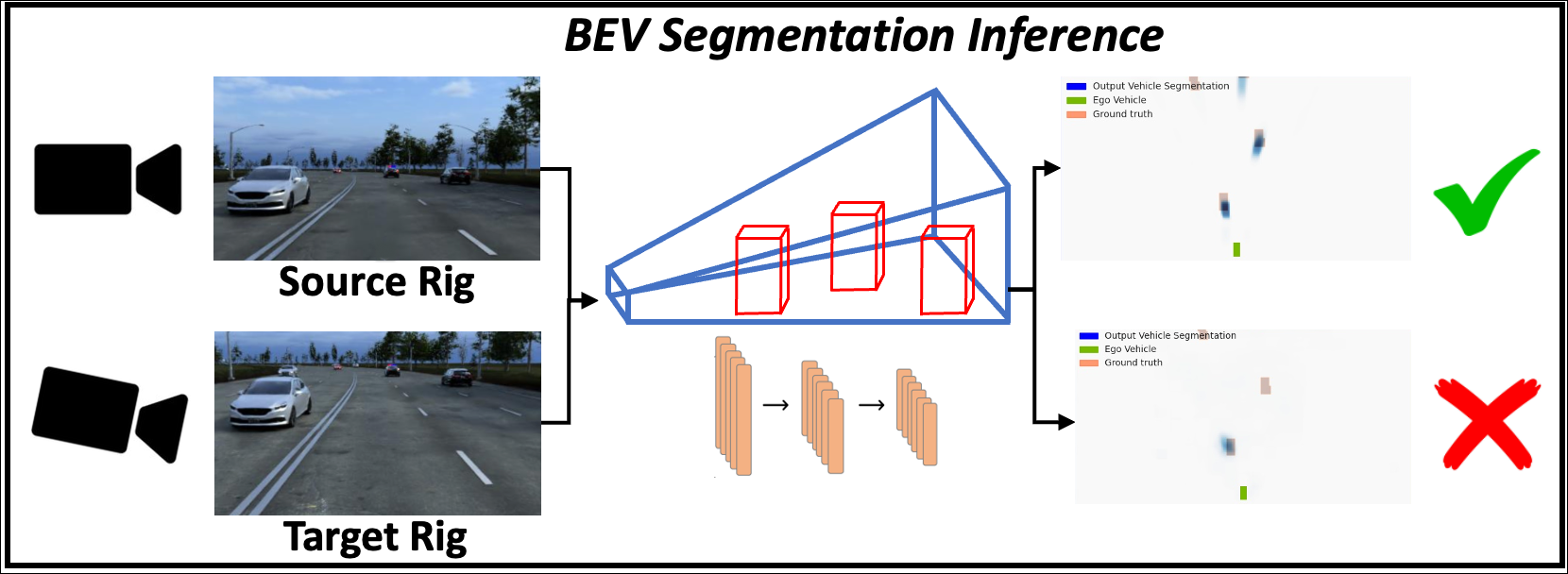}
        \caption{\textbf{Impact of Changed Camera Viewpoint:} We find that the performance of state-of-the-art methods for bird's eye view (BEV) segmentation quickly drop with small changes to viewpoint at inference. Above we see predictions from Cross View Transformers \cite{zhou2022cross} trained on data from a source rig (top). The target rig pitch is reduced by 10$^{\circ}$ (bottom), leading a 17\% drop in IoU.}
    \label{fig:overview}
    \vspace{-2mm}
\end{figure}


The goal of this paper is to bring understanding and a first approach to a real-world problem in the AV space that has yet to receive attention in the research literature -- generalization from a source to target camera rig. We focus on bird's eye view (BEV) segmentation from RGB data to motivate how changing camera viewpoint can affect AV perception models. We study this problem by conducting an in-depth analysis on the impact changing the camera viewpoint at inference time has on recent BEV segmentation models. Our findings indicate that even small changes in camera placement at inference time degrade BEV segmentation accuracy, as illustrated in Fig. \ref{fig:overview}. We then propose a method to improve generalization to a target rig by simulating views in the target perspective. We show that incorporating data generated from novel view synthesis into training can significantly reduce the viewpoint domain gap, bringing the BEV segmentation model to the same level of accuracy as when there is no change in camera viewpoint, without having to collect or label any additional data. We compare our approach with other strategies, such as augmenting the camera extrinsics and labels during training, and find that our approach leads to better accuracy. Little work has focused on the impact of viewpoint changes for AV perception, and, to the best of our knowledge, we are the first to study the impact of diverse camera viewpoint changes on 3D AV perception tasks, such as BEV segmentation. We hope that this paper will encourage more research on the important problem of \emph{viewpoint robustness} in AV.

\begin{figure*}
    \centering
    \includegraphics[scale=0.5]{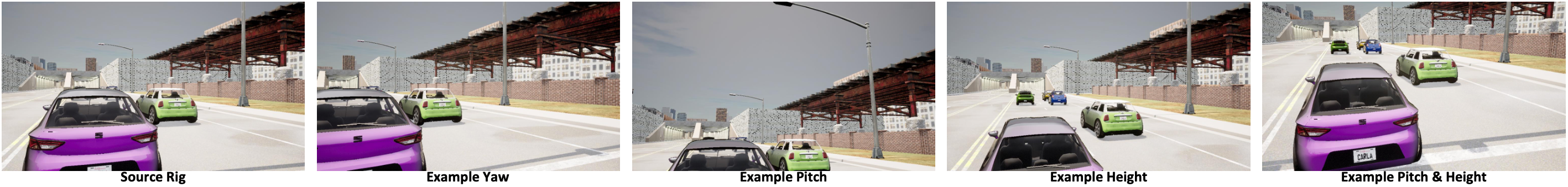}
        \caption{\textbf{Datasets rendered in CARLA across viewpoints:} For the analysis part of our work, we use CARLA to simulate different viewpoints. We rendered datasets from a total of 36 viewpoints, a few of which are highlighted above, including the source rig (extrinsics from nuScenes~\cite{caesar2020nuscenes} dataset), +12$^{\circ}$ yaw, +12$^{\circ}$ pitch, +21 inch height, and -12$^{\circ}$ pitch and +18 inch height together.}
    \label{fig:carla_examples}
\end{figure*}

\begin{figure*}
    \centering
    \includegraphics[scale=0.54]{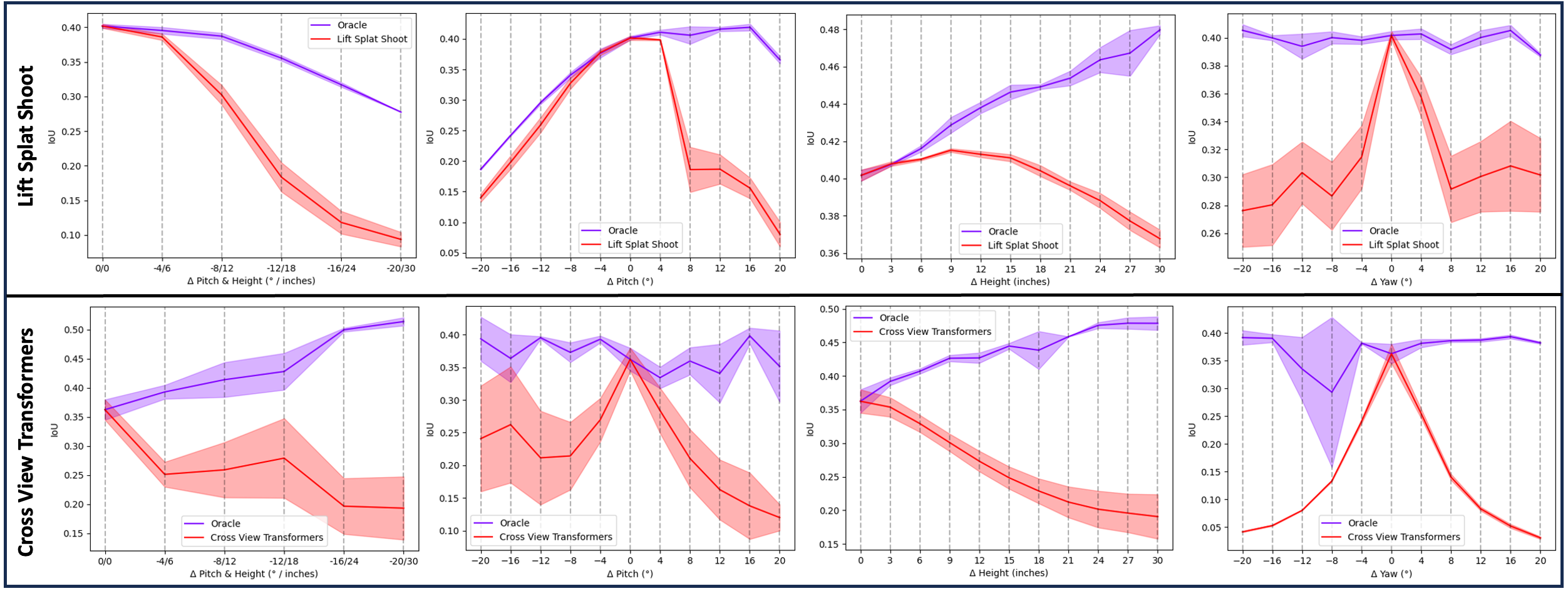}
        \caption{\textbf{Analysis of impact of viewpoint changes in CARLA:} We train a source BEV model using Lift Splat Shoot (LSS) \cite{philion2020lift} and Cross View Transformers (CVT) \cite{zhou2022cross}, denoted at point 0 on the $x$ axis of each graph. We then test the model across different target rigs where the camera pitch, yaw, height, or pitch and height are changed, as denoted by the different points along the $x$ axes. We also trained each model on the target rig directly and refer to this model as the "oracle", as it reflects the expected upper bound IoU for each viewpoint.}
    \label{fig:carla_analysis}
    \vspace{-2mm}
\end{figure*}

\vspace{2mm}
\noindent
Our paper makes the following contributions:

\begin{itemize}
\item We highlight the understudied problem of \emph{viewpoint robustness} in bird's eye view segmentation for autonomous vehicles (AV) through an in-depth analysis revealing that recent models fail to generalize to different camera viewpoints at inference time.


\item We propose a viewpoint augmentation framework for AV; we develop a novel view synthesis method that can be used to transform training data to target viewpoints and show that it improves the robustness of bird's eye view segmentation models to viewpoint changes.

\item We provide datasets that can be used to benchmark future work on viewpoint robustness in AV.

\end{itemize}
Because real-world AV datasets from a diverse set of camera rigs are not publicly available, we use simulated data both for (1) training and evaluation in our analysis and (2) evaluation of our proposed technique. Our synthetic datasets can be used for future efforts to benchmark the generalization abilities of different AV perception methods to viewpoint changes. Datasets are publicly available on our project page. The first dataset, rendered from CARLA, consists of both training and testing data, allowing for isolated analysis of the impact of viewpoint changes on BEV segmentation models (example images in Fig. \ref{fig:carla_examples}). The second dataset, rendered with NVIDIA DRIVE Sim~\cite{drivesim}, is significantly more photorealistic and consists of test sets from a diverse set of camera viewpoints. Thus, it can be used to evaluate models trained on real data, as we show in Section \ref{sec:results}. Both datasets include 3D bounding box labels.



\section{Related Work}

\subsection{Viewpoint Robustness}
Recent work has drawn attention to the susceptibility of NNs to misclassify when presented with distributions not seen during training. Madan \emph{et al.} \cite{madan2021small} show that both convolutional- and transformer-based classifiers are fooled by small viewpoint changes, and they introduce a search strategy for finding adversarial viewpoints, which leads to misclassifications over 71\% of the time. Similarly, \cite{madan2020and} shows that small viewpoint changes degrade classification performance, especially when paired with out of distribution (OOD) categories, and demonstrates that increasing the diversity of training data is an effective strategy to mitigate this issue. Do \emph{et al.} \cite{do2020surface} use homography to move images closer to the distribution of training data at inference time. Coors \emph{et al.} \cite{coors2019nova} study the impact of viewpoint changes for 2D semantic segmentation for AV, but do not explore 3D tasks. In contrast, we focus on providing a thorough analysis and a solution to the problem of viewpoint robustness for 3D AV perception tasks, focusing on BEV segmentation. 

\subsection{Novel View Synthesis}
Novel view synthesis (NVS) provides a way to render images from unseen viewpoints of a scene, and thus could be used to improve the robustness of perception models to viewpoint changes. Many methods have been proposed for NVS in recent years \cite{muller2022autorf,mildenhall2022nerf,niemeyer2022regnerf}, many of which are based on Neural Radiance Fields (NeRF) \cite{mildenhall2021nerf}. However, NeRF still faces two challenges that limit its applicability for our use case: (1) getting NeRF to generalize to dynamic scenes, which are common in AV, is an open research problem, and while there is promising work in this direction \cite{wu2022d,pumarola2021d}, the setup is often too constrained and simplified to fit the AV problem setting, and (2) NeRF is challenging to scale due to lack of generalizability, so multiple NeRFs must be trained to perform NVS across scenes. While there is work aimed at generalizing NeRF \cite{johari2022geonerf}, it remains an open problem and current methods are often constrained. Other methods for NVS rely on monocular depth estimation and can generalize across scenes when the depth estimation network is trained on diverse data. We leverage Worldsheet in our work \cite{hu2021worldsheet}, which is described in more detail in Sec. \ref{section:worldsheet}.


\subsection{Bird's Eye View Segmentation}
\label{sec:related_bev}
Bird's eye view (BEV) segmentation --- the task of segmenting a scene in the top-down view (BEV) from 2D images --- is a useful task for benchmarking AV perception \cite{philion2020lift,Roddick_2020_CVPR,acuna2021towards}. BEV segmentation requires a 2D to 3D unprojection to predict the position of objects surrounding the ego-vehicle from the BEV perspective. BEV segmentation models usually consist of an image encoder, which extracts the features from images from the camera rig, and a decoder, which uses the image features to predict the objects of interest in the BEV coordinate frame.  Existing methods condition on the extrinsics and intrinsics of each camera in different ways.  Lift-Splat-Shoot (LSS)~\cite{philion2020lift} and  Orthographic Feature Transform (OFT)~\cite{roddick2018orthographic} unproject features into a point cloud according to each camera's intrinsic and extrinsic parameters. LSS performs sum pooling along each pillar in the map-view, while OFT performs average pooling. Other methods, such as Cross View Transformers (CVT)~\cite{zhou2022cross}, treat camera intrinsics and extrinsics as a feature, rather than explicitly unprojecting. We use LSS and CVT to conduct benchmarks, since these two methods encompass both convolutional and transformer-based architectures and explicit and implicit geometric representations.

\section{Measuring the Impact of Camera Viewpoint Variations on BEV Segmentation}
\label{sec:analysis}

\noindent
\textbf{Method:} In this section, we introduce our approach and results for measuring the impact of changing the camera viewpoint at inference time for BEV segmentation models trained on a single, source rig. We use simulated data from CARLA~\cite{dosovitskiy2017carla} for this analysis for two reasons: (1) using simulated data allows us to isolate the domain gaps between training and testing such that only camera viewpoint changes, and (2) real AV datasets with large differences in camera position are not publicly available. Examples of different camera viewpoints rendered in CARLA are shown in Fig. \ref{fig:carla_examples}. For simplicity and ease of interpretation of our results, we conduct all experiments on a single camera rig, containing a front facing camera, which we refer to as the source rig. We first train a BEV segmentation model on data rendered from the source rig. For this rig, we use the camera parameters of sessions from the nuScenes dataset \cite{caesar2020nuscenes}. Then, we render train and test datasets from different target rigs, which contain variations to the yaw, pitch, height, or pitch and height of the camera. The train datasets are used to train an oracle for each target rig, while the test datasets are used to evaluate the model trained on the source rig in comparison to the oracle. For completeness, we sweep over a large range of each extrinsic and render a train and test dataset on regular intervals. For pitch and yaw, we sweep from -20$^{\circ}$ to 20$^{\circ}$, rendering a dataset every 4$^{\circ}$. For height, we sweep from 0 in to 30 in, rendering a dataset every 3 in. For height and pitch together, we sweep from 0$^{\circ}$ and 0 in to -20$^{\circ}$ and 30 in, rendering a dataset at every -4$^{\circ}$ and 6 in. 


To understand the domain gap introduced by changes in camera position, we test the ``source model'', which is the model trained on data from the source rig, across each test dataset, where each test dataset contains changes to either yaw, pitch, height, or pitch and height together. We then compare the test accuracy of the source model to the test accuracy of the oracle model, which was only trained on data from the target rig. The oracle model serves as an upper bound on model performance since there is no domain gap between the train and test datasets.




\noindent
\textbf{Model Details:} We conduct our analysis using two BEV segmentation models, Lift Splat Shoot (LSS) and Cross View Transformers (CVT). LSS uses an explicit geometric operation to map objects in the camera coordinate system to the bird's eye coordinate system. It does this by using each camera's intrinsic and extrinsic parameters to construct a frustum shaped point cloud per camera where predicted objects are placed inside. A convolutional encoder maps images to features and depths, which are unprojected into the frustum, and a cumulative summing operation is done over the features in the vertical pillars of the frustum before the decoder then predicts the final segmentation. In contrast, CVT uses a transformer to learn features over images, extrinsics, and intrinsics. The extrinsic and intrinsic parameters are used to condition the segmentation network, such that it implicitly learns correlations between the parameters and positions of objects relative to the ego-vehicle. We use these two architectures because they cover both explicit and implicit geometric representations and convolutional and transformer backbones, allowing us to test the impact of each on generalization to viewpoint changes.


\noindent
\textbf{Results:} Results of our analysis are shown in Fig. \ref{fig:carla_analysis}. We see that the performance of both LSS and CVT suffers drastically with even small changes to camera viewpoint, whether it be pitch, yaw, height, or pitch and height together. Because of the architecture of LSS, which includes cumulative summing in the vertical pillars within each frustum, changes to camera height have a relatively small impact on downstream BEV segmentation performance in comparison to other viewpoint changes. CVT lacks this generalization to changes in camera height because it does not sum features in the height dimension, but rather conditions on the camera extrinsics. We also note that because the training dataset is acquired in simulation, the extrinsics of the source rig have no noise or calibration error, and, thus, are always the same during training. As a result, we found that CVT learns to ignore the extrinsic embedding during training, indicating that the degradations we see to test performance in Fig. \ref{fig:carla_analysis} are the result of the images being out of distribution. In contrast, our experiments in Sec. \ref{sec:results} involve training CVT on real world data, which has calibration error, and, as a result, CVT learns to use the extrinsic embedding to inform predictions, but still lacks generalization to target rigs.

During our analysis, we also found that while changes to yaw have a negative impact on performance on LSS, the resulting segmentation predictions are transformed based on the difference in yaw between training and testing. To mitigate this, a post-processing step can be applied where the predictions are rotated to the viewpoint of the target rig. Post-processing can be used to mitigate the effect of changes in yaw, but does not generalize to other extrinsic parameters, such as pitch and height.

Lastly, we note two biases in the oracle models. First, we observe that the LSS oracle model trained on negative pitches performs poorly. Second, both LSS and CVT achieve higher test IoU when trained and tested with rigs with a larger camera height. While higher IoU could be explained by fewer occlusions due to a higher viewpoint, and thus more ground truth pixels, the number of ground truth objects is consistent across each of the test datasets (7 objects per frame on average), and so this bias is not explained by differences in the number of ground truth pixels. We note these biases, but they are not the main focus of our work. 


\vspace{1mm}
\noindent
\textbf{Training Details:} We train each BEV segmentation model three times and show the mean and standard deviation in test IoU in Fig. \ref{fig:carla_analysis}. Each model is trained on 25,000 images rendered from the front center camera (same camera parameters as in nuScenes) with the CARLA Simulator~\cite{dosovitskiy2017carla}. Train datasets are created for all camera viewpoints tested so that an oracle model can be constructed. For evaluation, we use 5,000 test images from each target rig, where the target rigs include changes to camera pitch, yaw, height, or pitch and height together, and are rendered from a different CARLA map than the training sets. Each model is trained for 30 epochs. We will release all 36 train and test datasets with this paper. The 36 datasets include train and test data for the source rig, 10 pitch rigs, 10 yaw rigs, 10 height rigs, and 5 height and pitch rigs.

\section{Viewpoint Robustness via NVS}

\begin{figure*}
    \centering
    \includegraphics[scale=0.27]{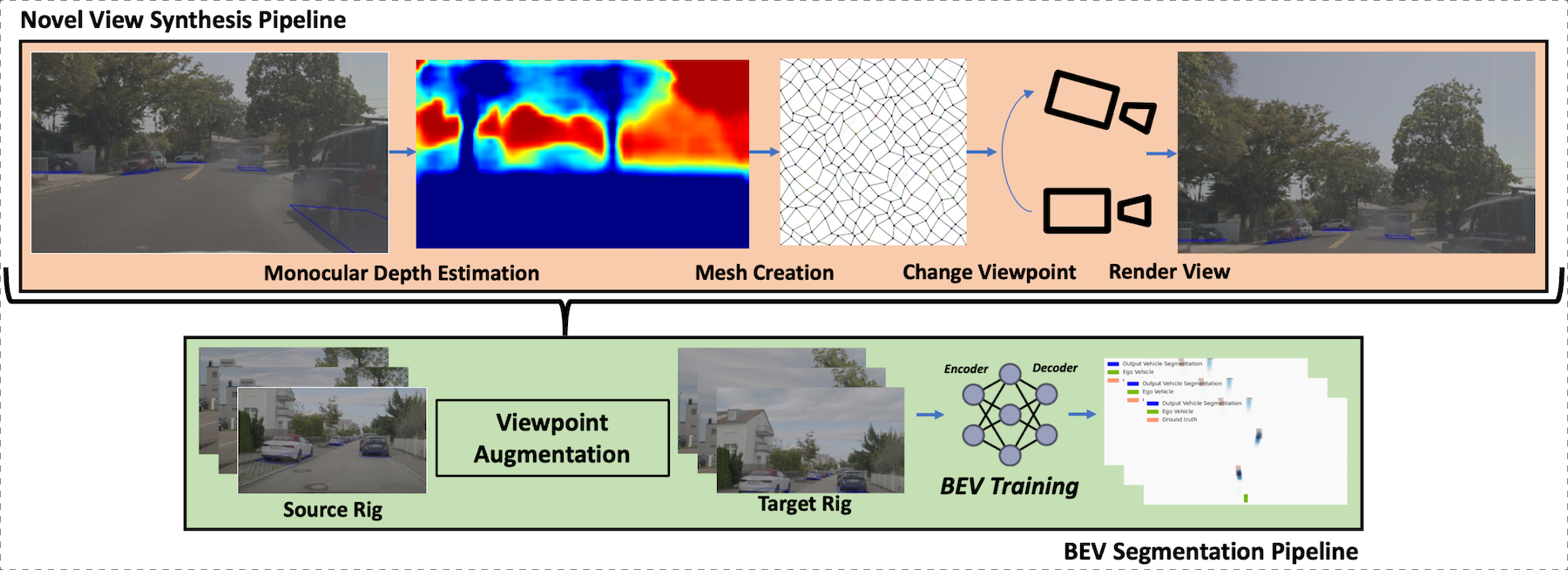}
        \caption{\textbf{Proposed Pipeline}. Current methods for bird's eye view (BEV) segmentation are trained on data captured from one set of camera rigs (the source rig). At inference time, these models perform well on that camera rig, but, according to our analysis, even small changes in camera viewpoint lead to large drops in BEV segmentation accuracy. Our solution is to use novel view synthesis to augment the training dataset. We find this simple solution drastically improves the robustness of BEV segmentation models to data from a target camera rig, even when no real data from the target rig is available during training.}
    \label{fig:method_overview}
\end{figure*}

We present a new method that improves generalization of BEV segmentation models to different camera positions using novel view synthesis (NVS). As described in Sec. \ref{sec:analysis}, BEV segmentation models fail to generalize to even small changes in camera viewpoint. However, collecting new data from each target rig, especially when AV companies may wish to deploy models across many types of cars, is impractical due to the cost of collection and annotation. Thus, we focus on NVS as it provides an opportunity to reuse labeled data from the source rig by transforming it into the viewpoint of each target rig. We can then train a new model on the transformed data for each target rig. We first define our NVS method. The key difference between our NVS method and past work is how we generalize to complex, dynamic AV scenes. Then, we show how the transformed data can be used to train BEV segmentation models for diverse target rigs without access to real data from the target rig. We use real data to train our NVS and BEV segmentation models. To evaluate over diverse target rigs, we use synthetic data rendered with NVIDIA DRIVE Sim since real data only provides one rig setting. We compare test performance achieved with models trained with data transformed to the target viewpoint vs. only data from a source rig. Our approach is summarized in Fig. \ref{fig:method_overview}.



\subsection{Preliminaries}
\label{section:worldsheet}

We build off of Worldsheet \cite{hu2021worldsheet}, a recent method for single image NVS of \emph{static scenes}, extending it to work on complex AV scenes that have \emph{dynamic} objects and occlusions. While NeRF-type approaches generate impressive NVS results, generalizing to dynamic scenes and across many scenes is still an active area of research. Worldsheet, on the other hand, is able to generalize across scenes, which is why we choose to use it in our work. The goal of Worldsheet is to build a 3D scene mesh, $M$, by warping a $W \times H$ lattice grid onto the scene based on predicted depths and vertex offsets. Given an input image, $I$, a ResNet-50 \cite{he2016deep} is trained to predict depth, $z$, and grid offset of each vertex, $V_{(x,y)}$ at each $(x,y)$ in $I$. $z$ and $V_{(x,y)}$ are used to build $M = (\{V_{(x,y)}\},\{F\})$, where $F$ are the mesh faces. A differentiable texture sampler is then used to splat the RGB pixel intensities from the original image onto the mesh's UV texture map. The pipeline is trained end-to-end on a multi-view consistency loss. Given two views of the scene, an input and a target, the mesh is predicted from the input view and then projected to the target view based on the target camera pose, $\theta_{t}$. The target view is rendered and compared to the GT with L1 and perceptual losses. A pix2pixHD generator inpaints parts of the scene in the generated target view that were not visible in the input. In contrast, we omit the pix2pixHD generator and use lidar depth supervision (LS), SSIM loss \cite{wang2004image}, automasking (AM) \& minimum loss (ML) over neighboring frames \cite{godard2019digging} to build an NVS model that generalizes to complex, dynamic, AV scenes.



\subsection{Novel View Synthesis for AV Data}


\begin{figure}[!t]
    \centering
    \includegraphics[width=\linewidth]{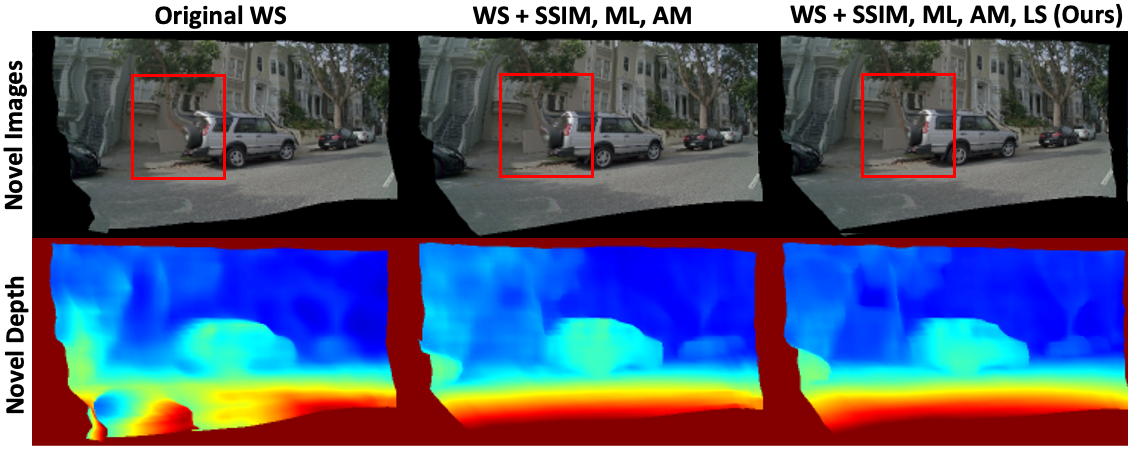}
        \caption{\textbf{NVS Qualitative Comparison:} We compare the unrectified NVS results (top) and depth results (bottom) from Worldsheet~\cite{hu2021worldsheet} (right) to our method (middle and left). SSIM is SSIM loss, ML is min loss, AM is automasking, LS is lidar supervision.}
        
    \label{fig:depth}
\end{figure}

\begin{figure*}
    \centering
    \includegraphics[scale=0.195]{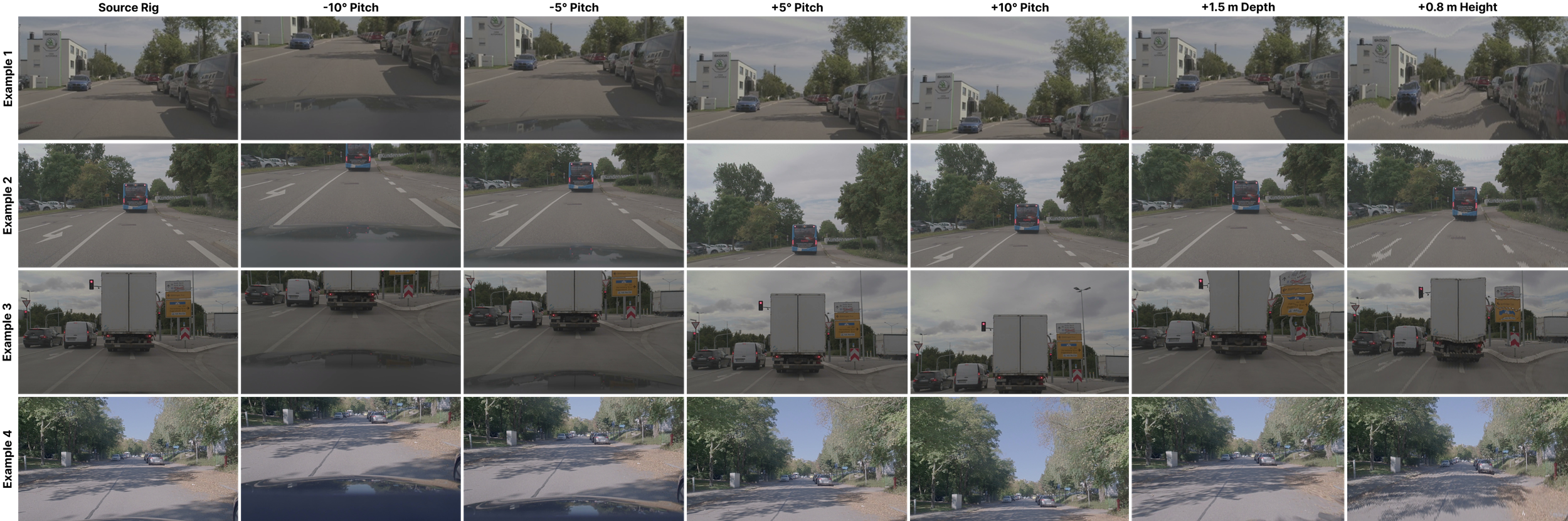}
    \caption{\textbf{Novel View Synthesis Qualitative Results:} Shown above are the novel view synthesis results (rectified) obtained with our method. We transform images from the source rig to each of the target viewpoints and then use them for BEV segmentation training.}
    \label{fig:nvs_results}
    \vspace{-0mm}
\end{figure*}


\noindent
\textbf{Overview:} Because AV sessions, $S$, are composed of temporally sequential images, $\{I_{0}, I_{1}, ... , I_{n}\} \in S$, temporal consistency, rather than multi-view consistency, can be enforced between neighboring images to train our NVS model, assuming a sufficiently high frame rate so parts of the scene are visible in the input and target images. For every input image, $I_{n}$, we enforce consistency between $I_{n-1}$ and $I_{n+1}$ by transforming $I_{n-1}$ and $I_{n+1}$ into the viewpoint of $I_{n}$ and comparing each predicted novel view $\hat{I_{n}}$ to GT $I_{n}$: 

\begin{align}
\label{eq:image}
\begin{gathered} 
\{\mathbf{\hat{I}}_{n}^{n+1},\mathbf{\hat{D}}_{n}^{n+1}\} = render(\{V_{(x,y)}^{n+1}\},\{F^{n+1}\}, T^{n+1}) \\
\{\mathbf{\hat{I}}_{n}^{n-1},\mathbf{\hat{D}}_{n}^{n-1}\} = render(\{V_{(x,y)}^{n-1}\},\{F^{n-1}\}, T^{n-1}) \\
L_{im} = \frac{1}{\pixelnum}\sum_{i=1}^{\pixelnum}\min(|I_{n,i} - \mathbf{\hat{I}}_{n,i}^{n+1}|, |I_{n,i} - \mathbf{\hat{I}}_{n,i}^{n-1}|)
\end{gathered} 
\end{align}

where $V$ are vertices, $F$ are mesh faces, and $T$ is the texture map. We render the meshes built from $I_{n-1}$ and $I_{n+1}$ in $I_{n}$'s viewpoint, forming novel view renderings $\hat{I_{n}}\in(\mathbf{\hat{I}}_{n}^{n+1},\mathbf{\hat{I}}_{n}^{n-1})$ and their corresponding depth maps $\mathbf{\hat{D}}_{n}^{n+1},\mathbf{\hat{D}}_{n}^{n-1}$. We then compute the per-pixel image loss $L_{im}$, where $\pixelnum$ is the valid pixel number and $I_{n,i}$ is the $i$-th pixel of $I_n$. 
Different from NeRF, worldsheet applies a single-layer mesh to synthesize novel views. In the discontinuous depth regions(\eg, boundaries), distortion might happen. 
To make the training more robust, we apply $L_1$ and SSIM loss between the GT image $I_n$ and the re-rendered image $\hat{I_{n}}$, where we follow the same setting in~\cite{godard2019digging}.

\vspace{1mm}
\noindent
\textbf{Occlusion Handling:} Inspired by unsupervised ego-video depth estimation work~\cite{godard2019digging}, we compute  two losses between $(I_{n}, \mathbf{\hat{I}}_{n}^{n-1})$ and $(I_{n}, \mathbf{\hat{I}}_{n}^{n+1})$, and pick up the minimal loss (ML) between them in a pixel-wise way. Intuitively, as the car is moving, some parts of the scene might be occluded in the last or next frame. However, they are less likely to be occluded in both two frames. Therefore, applying minimal losses  help prevent occlusions from affecting the training loss. 
We also use auto-masking~\cite{godard2019digging} to 
ignore pixels that violate camera motion assumptions, \eg, ego-car shadows.

\vspace{1mm}
\noindent
\textbf{Depth Supervision:}
Unlike other applications where only an RGB sensor is available, AVs are often equipped with lidar during data collection. We assume that lidar observations are available when training our NVS model. Thus, we can leverage lidar supervision (LS), rendering lidar into a ground truth sparse depth map~\cite{ravi2020pytorch3d}, $D_{n}$ for every image, $I_{n}$. To  further improve the quality of the lidar depth maps, we use two types of automasking (AM). First, we use a pre-trained sky segmentation network~\cite{tao2020hierarchical} to mask out the sky and set the depth for this part of each training image to infinity. Second, we use MaskRCNN~\cite{He_2017_ICCV} to predict masks of the ``close-by'' cars so that they are ignored in the depth loss, due to the fact that the lidar detector is mounted higher than the camera and it typically cannot see the close cars.

We then apply two depth losses, an L1 loss between the predicted depth and GT lidar depth (\emph{direct} depth loss) and the L1 loss between the predicted depth and ground truth depth after the prediction is projected into the viewpoint of the cameras at frame $n+1$ and $n-1$ (\emph{rendered} depth loss). As above, we also use minimal loss for depth supervision:

\begin{table}
\centering
\resizebox{1\columnwidth}{!}{%
\begin{tabular}{|l|cccc|}
\hline
Approach & Im. L1 $\downarrow$ & PSNR $\uparrow$ (dB) & SSIM $\uparrow$ & Depth L1 $\downarrow$\\
 \hline
   WS (original) & 0.145 & 22.602 & 0.595 & 0.00763 \\
   WS + SSIM, ML, AM & 0.141 & 22.819 & 0.606 & 0.00707 \\
   WS + SSIM, ML, AM + LS (Ours) & \textbf{0.138} & \textbf{22.936} & \textbf{0.608} & \textbf{0.00657} \\ 
   \hline
\end{tabular}
}
\vspace{0mm}
\caption{\textbf{NVS Ablation:} 
We ablate our changes, which improve NVS and depth over Worldsheet (WS). We test with 1K images.} \label{table:nvs_ablation}
\end{table}



\begin{equation}
\label{eq:depth}
\begin{gathered} 
L_{D}^{direct} = \frac{1}{\pixelnum}\sum_{i=1}^{\pixelnum}|D_{n-1,i} - F_{depth}(I_{n-1,i})| + \\|D_{n+1,i} - F_{depth}(I_{n+1,i})|\\
L_{D}^{rendered} = \frac{1}{\pixelnum}\sum_{i=1}^{\pixelnum}min(|D_{n,i} - \mathbf{\hat{D}}_{n,i}^{n+1}|, |D_{n,i} - \mathbf{\hat{D}}_{n,i}^{n-1}|)
\end{gathered} 
\end{equation}

Fig. \ref{fig:depth} shows how our method (SSIM, ML, AM, LS) improves depth estimation and NVS compared to Worldsheet. These improvements are quantitatively validated in Table \ref{table:nvs_ablation}.

\vspace{1mm}
\noindent
\textbf{Inpainting:} We train and test our NVS model using images from a 120$^{\circ}$ f-theta camera. The images are then rectified to 50$^{\circ}$ after NVS, such that missing parts of the scene not in the field of view of the final image. As a result, no image inpainting is needed. 
Our NVS results are shown in Fig. \ref{fig:nvs_results}.

\begin{figure}
    \centering
    \includegraphics[scale=0.36]{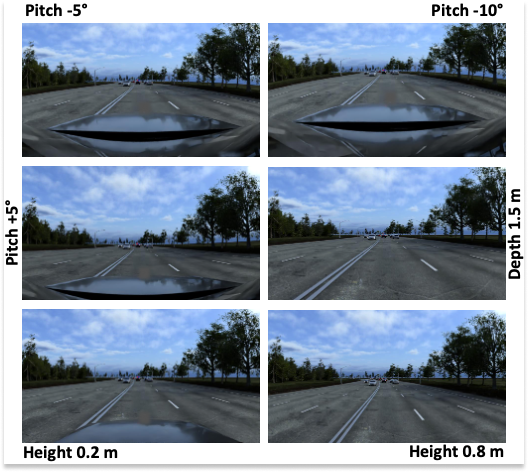}
        \caption{\textbf{Evaluation Data:} We use images from NVIDIA DRIVE Sim \cite{drivesim} to evaluate our method on a diverse set of target rigs. Shown here are example test images with different viewpoints.}
    \label{fig:evaluation_images}
\end{figure}

\subsection{Augmenting BEV Segmentation Training}

The focus of our paper is not on NVS quality, but on the impact using NVS generated data can have on the problem of \emph{viewpoint robustness} in AV. Given a labeled BEV segmentation training dataset, $D_{source}$, of size $N$, we use our NVS method to transform $n$ images from $D_{source}$ to the viewpoint of the target rig, obtaining $D_{target}^{pred}$ of size $n$. This transformation is done by (1) estimating the depth each image, (2) creating meshes, (3) changing the viewpoint of the cameras, and 4) rendering each image in the viewpoint of the target rig. Finally, we construct a new BEV dataset, $D_{final}$ of size $N$, containing the $n$ transformed images from $D_{target}^{pred}$ and $N-n$ images from $D_{source}$. The number of transformed images, $n$, is a hyperparameter and in our experiments we transform 25\%, 50\%, or 100\% of $D_{source}$ to the viewpoint of the target. The reason we do not always transform all $N$ images is the NVS model may introduce other domain gaps; an ablation on this is done in Sec. \ref{sec:discussion}. We train both the NVS model and BEV segmentation model on a real-world dataset, described in Sec. \ref{sec:datasets}. An overview of the training pipeline is shown in Fig. \ref{fig:method_overview}.

\section{Experiments and Results}
\label{sec:results}

We show the effectiveness of our method by using it to train BEV segmentation models for diverse target rigs, without any access to real data from the target rig during training. We first train our NVS model to transform data from the source rig to the target viewpoint. Next, we transform some or all of the source rig training data to the target rig. Finally, we train the BEV segmentation model for the target rig using a combination of transformed data and source data. All training is done on real world data, but evaluation is done with NVIDIA DRIVE Sim, allowing us to test across target rigs that are not available in public datasets. 



\begin{table}
\small
\centering
\begin{tabular}{p{0.13\linewidth}p{0.15\linewidth}p{0.10\linewidth}p{0.10\linewidth}p{0.14\linewidth}p{0.07\linewidth}}
\hline
 Extrinsic & $\Delta$ View & Source & Source* & Extr Aug & Ours \\
\hline
\hline
- & 0 & 0.170 & 0.170 & 0.155 & - \\ 
\hline
Pitch & -10$^{\circ}$ & 0.014 & 0.078 & 0.126 & \textbf{0.165} \\
Pitch & -5$^{\circ}$ & 0.037 & 0.141 & 0.128 & \textbf{0.161} \\
Pitch & +5$^{\circ}$ & 0.016 & 0.076 & 0.028 & \textbf{0.173} \\ 
Depth & 1.5 m & 0.017 & 0.156 & 0.150 & \textbf{0.174} \\ 
Height & 0.2 m & 0.094 & 0.175 & 0.145 & \textbf{0.177} \\
Height & 0.8 m & 0.003 & 0.170 & 0.132 & \textbf{0.214} \\
\hline
\end{tabular}
\caption{\textbf{Results:} We report the IoU of the CVT model trained on a source rig and tested across target rigs where pitch, depth, and height are changed (source). We then compare against two baselines, described in text. Last, we compare with our method, which is trained with some data transformed to the target rig view. The first row shows IoU of the source evaluated on sim data from the same viewpoint, and is our best estimate of oracle performance.}
\label{table:results}
\end{table}

\subsection{Datasets}
\label{sec:datasets}
\noindent
\textbf{Training:} We train both the NVS and the BEV segmentation model on an internal dataset of 43 real AV sessions. We subsample the images from each video at a higher frame rate for our NVS training dataset than our BEV segmentation training dataset, yielding 250,000 and 30,000 training images respectively. All images are captured from a 120$^{\circ}$ f-theta lens camera. Prior to BEV segmentation training, we rectify the images to 50$^{\circ}$. Examples of rectified images from the source rig are shown in the first column of Fig. \ref{fig:nvs_results}.



\vspace{1mm}
\noindent
\textbf{Evaluation:} We use simulated data from challenging scenes for the evaluation since real datasets with large viewpoint changes are not available and collecting them across many views is impractical. Simulated data could be used for train and test, but generating sufficiently large and diverse simulated train datasets is difficult. To mitigate the domain gap of training on real data and testing on simulated data, we use NVIDIA DRIVE Sim. Example images are shown in Fig. \ref{fig:evaluation_images}. To measure the domain gap, we trained a model on real data and evaluated it on both a real test dataset and a simulated test dataset from the source rig. The gap was 7.5\% IoU, which is acceptable for our work, since we are concerned with relative changes in IoU, not absolute IoU.


\subsection{Experiment Details}
We demonstrate our method by transforming the dataset from the source rig, $D_{source}$, to the viewpoint of six target rigs, training a BEV segmentation model for each, and evaluating the model on simulated data from the target rig. We conduct experiments with a single camera rig. The target rigs include pitch -10$^{\circ}$, -5$^{\circ}$, and 5$^{\circ}$, depth 1.5 m, and height 0.2 m and 0.8 m. Examples of source rig data transformed to each of the target rigs with the NVS model are shown in Fig. \ref{fig:nvs_results}. We note that, quantitatively, the NVS quality is best for changes in pitch and lowest for large changes in height. Despite lower quality for some transformed viewpoints, we show that the transformed data still leads to significant improvements in BEV segmentation accuracy for each target rig. For each target rig, we train a Cross View Transformers (CVT) model three times, with 25\%, 50\%, and 100\% of $D_{source}$ transformed to the target rig viewpoint. We also train CVT on source rig data for comparison.



\subsection{Baselines}

\noindent
We compare against two baseline approaches:

\vspace{1mm}
\noindent
\textbf{ -- Using Train Extrinsics at Inference Time (\emph{Source*}):} By passing in the train extrinsics to the BEV segmentation model at inference time, we find that, despite the image itself being from a different rig, performance improves. 

\vspace{0mm}
\noindent
\textbf{ -- Extrinsic Augmentations (\emph{Extr. Aug.}):} Rather than augmenting the training images to be from the viewpoint of the target rig, we instead apply random rotations to both the extrinsic matrix and 3D bounding box labels together within the bounds of extrinsics of the target rigs. 

\begin{figure}[!t]
    \centering
    \includegraphics[scale=0.5]{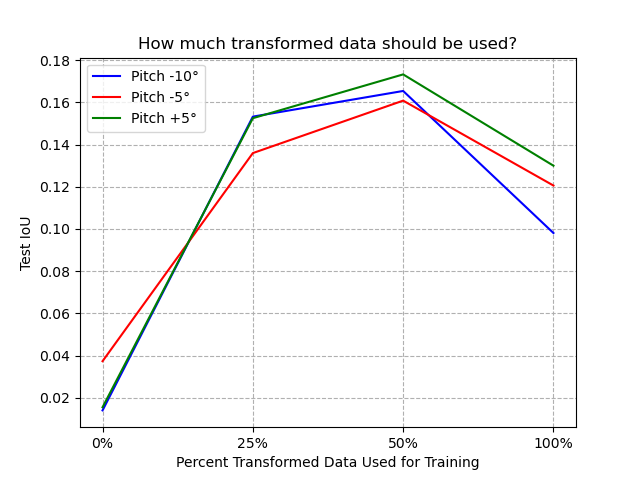}
        \caption{\textbf{Ablation: Varying percent transformed training data:} We observe that transforming 25-50\% of the training dataset to the viewpoint of the target rig results in the best test IoU.}
    \label{fig:ws_result_graph}
    \vspace{-2mm}
\end{figure}

\subsection{Results}
We find that our approach of training BEV segmentation models with 25\%, 50\%, or 100\% data transformed into the view of the target rig significantly improves BEV segmentation accuracy compared to training with only data from the source rig, leading to the same level of accuracy as when there is no viewpoint change. Results are shown in Table \ref{table:results}. We report the best IoU from the models trained with 25\%, 50\%, or 100\% transformed data, but note that all top performing models use only 25\% or 50\% transformed data, and the rest of the training data remains from the source rig. We observe that both baselines also significantly improve the IoU compared to the model trained only on source data, but not as much as our NVS approach. We also compute the IoU of the model trained only on the source rig and tested on synthetic data from the same viewpoint to serve as a reference upper bound for expected performance when there is no viewpoint gap, shown in the first row of Table \ref{table:results}. This upper bound is more reliable than training and testing on simulated data, which results in an average of 35.4\% IoU across views due to the lack of domain gap and limited diversity, resulting in visually similar train and test data. 



 Lastly, we conducted an experiment in which we trained a model on $\frac{1}{2}$ source rig and $\frac{1}{6}$ +5$^{\circ}$ pitch, $\frac{1}{6}$ +1.5 m depth, and $\frac{1}{6}$ +0.2 m height data, resulting in 0.19 mean test IoU across views in Tab. \ref{table:results} (0.206 for train views and 0.178 for other views). This result suggests training on multiple views can improve IoU over training only on the target view. Altogether, our results support our hypothesis that using NVS to transform labeled train data from the viewpoint of a source rig to that of a target rig and then training a BEV segmentation model with that data can enable the creation of BEV segmentation models for target rigs without the associated cost of collecting and annotating data from each target rig.




\section{Discussion}
\label{sec:discussion}

We observe that, despite some NVS transformations leading to artifacts, e.g. the +0.8 m height transformation, the images still significantly help downstream BEV segmentation models to generalize to the desired target rig. In addition to our main results, we also conduct two ablation studies on our method, which are described below.



\vspace{1mm}
\noindent
\textbf{Amount of Transformed Data:} An open question is how much data from the source rig dataset should be transformed to the viewpoint of the target rig. While transforming all of the data may lead to a content gap due to NVS being imperfect, transforming too little may not expose the BEV segmentation model to enough examples of data from the target rig viewpoint. In our experiments, we train BEV segmentation models with 25\%, 50\%, and 100\% transformed data. Shown in Fig. \ref{fig:ws_result_graph} is the IoU as a function of the amount of transformed training data. We see that IoU consistently increases as more transformed data is added to training until 50\%. The model trained with 100\% underperforms, most likely due to other domain gaps introduced by NVS. 

\vspace{1mm}
\noindent
\textbf{Interpolation and Extrapolation:} In our work, we focus on generating target rig specific BEV segmentation models without the cost of data collection. However, one may wish to create a single BEV segmentation model that generalizes to multiple camera rigs. We investigate whether our approach can enable that by testing how models trained with two viewpoints interpolate between those viewpoints and extrapolate beyond those viewpoints. We test all combinations of the pitch models trained with 50\% transformed data and 50\% source rig data, averaging test performance for interpolatation and extrapolation. An example of interpolation is testing a model trained on 0$^{\circ}$ and -10$^{\circ}$ pitch on -5$^{\circ}$ pitch, while an example of extrapolation is testing a model trained on 0$^{\circ}$ and -5$^{\circ}$ pitch on -10$^{\circ}$ pitch. On average, we find interpolation performance is 14.9\% IoU and extrapolation performance is 14.8\% IoU, suggesting the proposed method can improve generalization beyond the target rig.

\section{Conclusion}

We find that changing camera viewpoint, even by small amounts, has a significant impact on BEV segmentation models that have not been trained on that viewpoint. As AVs become more ubiquitous and companies scale across different vehicle types, this problem, which we dub \emph{viewpoint robustness}, will become critical to address. Our work makes a first attempt at improving viewpoint robustness using data generated from our method for NVS. We find that augmenting the BEV segmentation train dataset with data generated from the viewpoint of the target camera rig improves generalization to the target rig. As part of our work, we propose a method for NVS and show that it can be used to effectively mitigate the viewpoint domain gap.

\vspace{2mm}
\noindent
\textbf{Acknowledgements:} We thank Alperen Degirmenci for his valuable help with AV data preparation and Maying Shen for her valuable support with experiments.

{\small
\bibliographystyle{ieee_fullname}
\bibliography{egbib}
}

\end{document}